\documentclass[10pt,twocolumn,letterpaper]{article}

\usepackage{cvpr}              
\usepackage[accsupp]{axessibility}

\usepackage[dvipsnames]{xcolor}

\usepackage{comment}

\newcommand{\yin}[1]{\textcolor{blue}{[Yin: #1]}}
\usepackage[nice]{nicefrac}
\usepackage{multibib}
\newcites{Main}{References}
\newcites{Supp}{Supplementary References}

\definecolor{cvprblue}{rgb}{0.21,0.49,0.74}
\usepackage[pagebackref,breaklinks,colorlinks,citecolor=cvprblue]{hyperref}

\def\paperID{13093} 
\def\confName{CVPR}
\def\confYear{2024}

\title{Towards 3D Vision with Low-Cost Single-Photon Cameras}

\author{Fangzhou Mu$^\dagger$, Carter Sifferman$^\dagger$, Sacha Jungerman, Yiquan Li$^\ddagger$, Mark Han$^\ddagger$, \\ Michael Gleicher, Mohit Gupta, Yin Li\\[0.02in]
{\small $^\dagger$ co-first author, $^\ddagger$ equal contribution}\\[0.02in]
University of Wisconsin-Madison
}

\begin{document}
\twocolumn[{
    \renewcommand\twocolumn[1][]{#1}
    \vspace{-1em}
    \maketitle
    \vspace{-3em}
    \begin{center}
        \centering
        \includegraphics[width=0.95\textwidth]{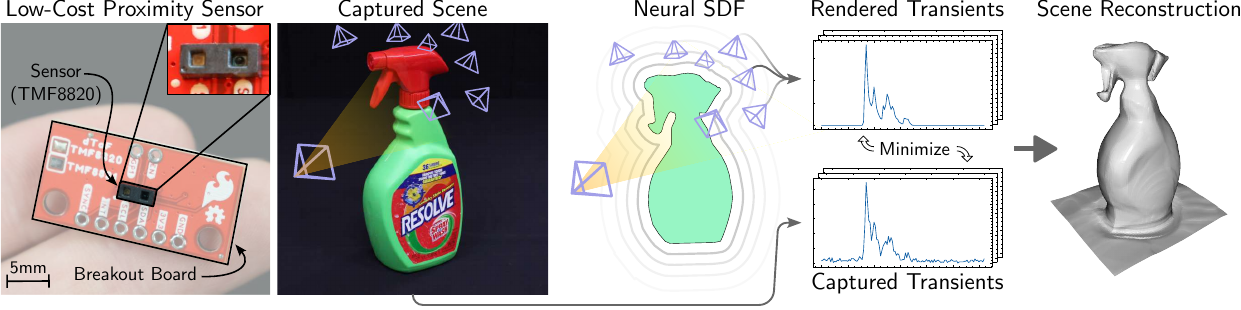}
        \vspace{-0.5em}
        \captionof{figure}{We demonstrate that measurements from spatially distributed low-cost single-photon proximity sensors (left) can be used to reconstruct 3D shape of real world objects (right). Our method combines a differentiable image formation model and neural rendering to recover 3D geometry based on measurements (transient histograms) from sensors with known poses. This is done by minimizing the difference between the observed and rendered sensor measurements. For clarity, a subset of sensor poses and measurements are shown. 
        }
        \label{fig:teaser}
    \end{center}
}]
\begin{abstract}
We present a method for reconstructing 3D shape of arbitrary Lambertian objects based on measurements by miniature, energy-efficient, low-cost single-photon cameras. These cameras, operating as time resolved image sensors, illuminate the scene with a very fast pulse of diffuse light and record the shape of that pulse as it returns back from the scene at a high temporal resolution. We propose to model this image formation process, account for its non-idealities, and adapt neural rendering to reconstruct 3D geometry from a set of spatially distributed sensors with known poses. We show that our approach can successfully recover complex 3D shapes from simulated data. We further demonstrate 3D object reconstruction from real-world captures, utilizing measurements from a commodity proximity sensor. Our work draws a connection between image-based modeling and active range scanning, and offers a step towards 3D vision with single-photon cameras. Our project webpage is at \url{https://cpsiff.github.io/towards_3d_vision/}.
\end{abstract}\vspace{-1em}    
\section{Introduction}
\label{sec:intro}


Reconstructing 3D shape of real objects remains a central problem in vision, solutions to which have evolved into two parallel branches. \textit{Image-based modeling}~\citeMain{szeliski2022cv} leverages a plethora of visual cues from multiple photographs (\eg, stereo, motion, shading), leading to problems including multi-view stereo~\citeMain{seitz2006mvs}, photometric stereo~\citeMain{ackermann2015ps} and the more recent neural radiance fields (NeRF)~\citeMain{Mildenhall2020nerf}. Conversely, \textit{active range scanning}~\citeMain{jarvis1983range} combines an active light source with an imaging sensor, giving rise to imaging techniques such as structured light~\citeMain{geng2011sl}, and time-of-flight~\citeMain{hansard2012tof}. Conventional wisdom suggests that range scanning yields more precise 3D geometry than image-based modeling at the cost of using specialized, expensive hardware. 


An emerging approach for range scanning is direct time-of-flight imaging with active \emph{single-photon cameras}, a form of time-resolved image sensor. This approach couples a pico-to-nanosecond detector with a fast coherent light source, illuminates the scene with a very short pulse of light, and measures the intensity of the light over time as it reflects back from the scene. The resulting incident wavefront is recorded and quantized, forming a \textit{transient histogram}. A special case of this approach is single-photon LiDAR, in which the light source (laser) is highly focused, the detector finds the peak in the histogram, and the sensor reports a single distance value per detector pixel. When using a diffuse light source, these time-resolved sensors capture visual information beyond the distance measurements extracted in LiDAR. Transient histograms in this case record distributions of times-of-flight, encoding the product of scene geometry and reflectance over each imaged scene patch~\citeMain{Jungerman_ECCV_22}. 


The entirety of information in the transient histogram has been previously utilized in applications such as fluorescence lifetime imaging~\citeMain{lagarto2020real} and non-line-of-sight imaging~\citeMain{velten2012recovering}. Recently, time-resolved measurements from such sensors have been explored for line-of-sight 3D reconstruction and novel view synthesis~\citeMain{Malik2023transient}. Unfortunately, these systems, due to their prototype-like nature, exist in a bulky benchtop form factor, require significant power, and are costly (over \$10,000 USD). They are thus not suited for many applications including autonomous drones, wearable computing, and augmented reality, in which low-cost, energy efficient, miniature sensors are required. 

Low-cost single-photon cameras have recently become available in the form of active single photon avalanche diodes (SPADs). They include one or more SPADs paired with an eye-safe diffuse light source (\eg, an infrared VCSEL laser), are very small ($<$20 mm$^3$), inexpensive ($<$\$5 USD), and power efficient ($<$10 milliwatts per measurement)~\citeMain{TMF8820, VL6180X}. These are sold as proximity sensors and some can be configured to report transient histograms. Compared to laboratory-grade systems, however, they lack precise optics, calibration, and timing characteristics and have an order of magnitude lower spatial and temporal resolution. Still, these sensors have proven successful for material classification~\citeMain{Becker2023plastic}, human pose recognition~\citeMain{Ruget2022pixels2pose}, and simple shape recovery (\ie, a planar surface)~\citeMain{Jungerman_ECCV_22,Sifferman2023unlocking}.

In this work, we address the problem of reconstructing 3D shape of arbitrary Lambertian objects from a set of spatially distributed low-cost single-photon cameras with known poses. We present an approach that combines a neural signed distance field surface representation, a differentiable transient formation model for practical active single-photon cameras, and an optimization scheme following the analysis-by-synthesis pipeline. 
Fig.\ \ref{fig:teaser} illustrates our sensor, imaging setting, and approach. We show that our approach can successfully recover complex 3D shapes from simulated data. We further demonstrate 3D object reconstruction from real-world captures, utilizing measurements from a low-cost, off-the-shelf proximity sensor. 

Our approach draws a connection between image-based modeling and active range scanning while avoiding some of their pitfalls. By using an active light source and SPADs, our approach illuminates and images an object from multiple angles, resembling the imaging setup of multiview photometric stereo. Our model also echoes the design of NeRF. The key difference is that our model considers an input of low spatial resolution (\ie, hundreds of pixels per scene), and leverages rich temporal information encoded in the transient histograms. Unlike image-based modeling, our approach can operate under low light conditions, and reconstruct objects without texture. Unlike most range scanning methods, our approach performs favorably in the presence of strong ambient flux and is robust to mildly specular surfaces (see top of the spray bottle in Fig.\ \ref{fig:teaser}). 







\smallskip
\noindent \textbf{Practical Applications.} Our approach presents a compact, energy-efficient, and low-cost solution for 3D sensing when relative sensor poses are known. We envision applying our method to settings where relative sensor poses are fixed, and 3D geometry can be estimated on a per-frame basis by combining measurements from a distributed set of sensors. This includes applications like wearable hand tracking~\citeMain{devrio2022discoband,hu2020fingertrak}, and collision avoidance and mapping for drones, mobile robots, and robot manipulators. This work provides a first step towards enabling such applications.

\smallskip
\noindent \textbf{Scope and limitation.} Similarly to other methods for active range scanning, the proposed method fails on highly specular objects and requires hundreds of views to reconstruct a complex 3D shape. Our imaging system is tailored for low-cost sensors with limited range (4-6 meters) and temporal resolution (over 100 picoseconds), and thus it is unproven at imaging larger space or finer details. While methods exist for fast NeRF~\citeMain{muller2022instant}, our approach is not optimized for speed and takes a few hours to reconstruct a single object. Instead, we focus on demonstrating the feasibility of reconstructing real-world 3D objects using commodity sensors.

\section{Related Work}
\label{sec:related_work}


\noindent\textbf{Time-resolved imaging.} Time-resolved sensors have long been used in applications such as non-line-of-sight (NLOS) imaging~\citeMain{xin2019theory,Faccio2020non}, where a scene is recovered from around the corner, fluorescence lifetime imaging~\citeMain{lee2023caspi}, a microscopy technique for characterizing a biological sample, and to measure distance using direct time-of-flight~\citeMain{heide2018sub,gupta2019photon}. Recent techniques~\citeMain{Malik2023transient} used high-end time-resolved sensors to directly recover scene geometry. However, these methods rely on hardware prototypes which are often prohibitively expensive and not accessible to consumers. For example, Transient NeRF~\citeMain{Malik2023transient} performs view synthesis and 3D reconstruction using a lab prototype with no ambient light and $2$-$5$ views of $512 \times 512$ transient histograms, each with 1500 bins. In contrast, in this work, we use commodity hardware with ambient light and $128$-$240$ total transients, each comprised of $128$ bins.

\smallskip
\noindent\textbf{3D Imaging with low-cost SPADs.} With cheap time-resolved proximity sensors becoming commonplace, recent works have investigated their use for 3D reconstruction. Callenberg \etal~\citeMain{Callenberg2021CheapSPAD} demonstrate that, with some additional hardware, high-resolution depth imaging from a single viewpoint is possible. A low-cost SPAD has been used to augment an RGB SLAM system~\citeMain{tofslam}. Other works utilize supervised machine learning to recover geometric information from single low-cost SPAD measurements, such as 3D human pose \citeMain{Ruget2022pixels2pose} or high-resolution depth images \citeMain{Jungerman_ECCV_22,deltar}. Jungerman \etal~\citeMain{Jungerman_ECCV_22} use differentiable rendering to recover two degrees of freedom of a planar surface from a single low-cost SPAD transient histogram. Sifferman \etal~\citeMain{Sifferman2023unlocking} extend this method to fully recover a planar surface using a low-cost SPAD with multiple detector pixels.

\smallskip
\noindent\textbf{Neural implicit representations.} Neural representations, as popularized by NeRF~\citeMain{Mildenhall2020nerf}, enable novel view synthesis and 3D reconstruction by representing the scene as a neural network. While the original NeRF representation encoded view-dependent volumetric effects, alternative encodings have been proposed to better model geometry and reconstruct surfaces. NeuS~\citeMain{Wang2021neus} represents the scene as a level set, allowing for better modeling of surfaces at the expense of not being able to represent volumetric effects. Many works extend these ideas to work with different sensing modalities and external supervision, such as depth queues from structure-from-motion~\citeMain{deng2022dsnerf}, RGB images plus continuous-wave time-of-flight sensors~\citeMain{attal2021torf}, only depth information~\citeMain{OrtiziSDF2022,liu2023multimodal}, or more recently using only transients~\citeMain{Huang2023nfl,Malik2023transient}. In this work, we perform 3D reconstruction using only transient histograms as captured using commodity hardware by adapting the implicit surface representation introduced by NeuS to render transients. 

\smallskip
\noindent\textbf{Multiview photometric stereo.} Our approach illuminates and images an object from multiple views in order to reconstruct its 3D shape, having conceptual similarity to image-based modeling methods. This concept has previously been explored as multiview photometric stereo (MVPS)~\citeMain{hernandez2008mvps,cheng2021multi}: reconstructing 3D geometry given distributed views of the scene and various lighting conditions. More recent approaches resort to deep learning models~\citeMain{zhao2023mvpsnet} and consider NeRF-based representations~\citeMain{kaya2023multi,yang2022psnerf}.




\section{3D Vision with Single-Photon Cameras}
\label{sec:method}

\begin{figure*}[!ht]
    \centering
    \includegraphics[width=1.0\textwidth]{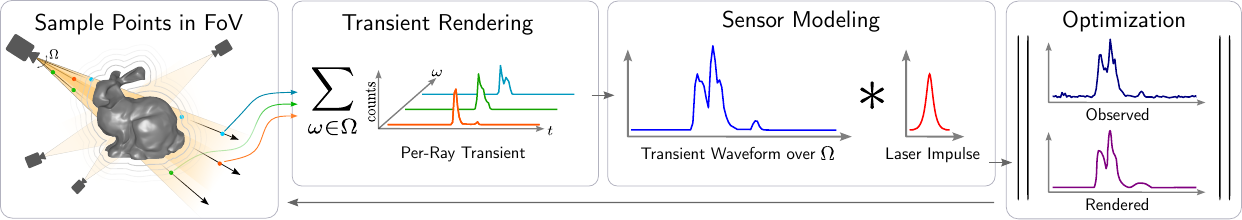}\vspace{-0.5em}
    \caption{\textbf{Method Overview:} The scene is modeled as a neural implicit surface in the form of an SDF. To render a transient, we approximate Eq.~\ref{eq:angular} by sampling rays within each pixel's FoV, and subsequently points on those rays. This idealized transient waveform is then convolved with the sensor's laser impulse response to model the transient histogram formation. Finally, we optimize the scene representation by minimizing a loss between the rendered transients and the observations. 
    }\vspace{-1.5em}\label{fig:method}
\end{figure*}

We propose to reconstruct 3D scene geometry using a sparse set of measurements from low-cost time-resolved SPAD sensors. 
Our approach assumes a distributed set of $N$ SPAD sensors with known pose, each comprising a single-pixel detector co-located with a diffuse laser. 
The time of flight, $t$, of returning photons, received over the detector FoV~\footnote{We assume that the illumination and detector FoV coincide and refer to both as the sensor FoV in the rest of the paper.} is recorded and binned into \emph{transient histograms}~\citeMain{Jungerman_ECCV_22,Sifferman2023unlocking}, $\{\mathbf{h}_j\}_{j=1}^N$: time-resolved measurements that encode rich information about the scene geometry and reflectance. Our goal is to recover scene geometry from these measurements. 

\smallskip
\noindent\textbf{Traditional depth ranging.} Our setup differs from the conventional use of SPADs for depth ranging~\citeMain{heide2018sub,gupta2019photon,Malik2023transient}, which assumes a collimated laser and focused detector. There, the depths of individual scene points can be independently estimated from separate transients, and a dense scan of thousands of scene points is needed to obtain full scene geometry. These systems require high-quality beam lasers and detectors with high temporal resolution. In contrast, our imaging configuration is indicative of the real-world use cases of low-cost SPAD sensors with \textit{diffuse lasers}, wide FoVs, and low-resolution detectors. In our case, the transient encodes rich information beyond its peak location. 

\smallskip
\noindent\textbf{Low-cost SPAD sensing.} Our setup extends recent systems~\citeMain{Jungerman_ECCV_22,Sifferman2023unlocking} built around low-cost SPAD sensors for 3D sensing in two ways. First, our system consists of multiple posed sensors as opposed to a single sensor. Second, it is capable of capturing and reconstructing complex, non-parametric scenes as opposed to the parametric geometry of a single plane. With these upgrades, our system is reminiscent of multi-view stereo and NeRF-like systems for conventional RGB cameras, and is a substantial step towards practical 3D vision using commodity proximity sensors.

\subsection{Transient Formation Model}\label{sec:model}

We model the transient formation process in two steps. We first derive the \emph{transient waveform} of a scene by modeling the interaction of light with scene geometry. The next step uses this waveform as the input to a sensor model that accounts for non-linearities that occur when capturing \emph{transient histograms} with real hardware.

\smallskip
\noindent\textbf{Transient waveform.}
As in~\citeMain{Jungerman_ECCV_22,Sifferman2023unlocking}, we ignore high-order light paths and only model direct reflection from the object surface. The transient waveform $\tau_{\mathbf{s}}(t)$ of a sensor $\mathbf{s}$ located at the origin can be written in path integral form~\citeMain{veach1998robust} as
\begin{equation}\label{eq:path}
\begin{split}
    \tau_{\mathbf{s}}(t)&=\int_{\mathcal{S}}\rho(\mathbf{x})\frac{G(\mathbf{x},\boldsymbol{\omega})}{\|\mathbf{x}\|^4}\delta\left(\|\mathbf{x}\|_2-\frac{ct}{2}\right)\mathrm{d}A(\mathbf{x}),\\
    G(\mathbf{x},\boldsymbol{\omega})&=f_r(\mathbf{x},\mathbf{n}_{\mathbf{x}},\boldsymbol{\omega}) V(\mathbf{x})\langle\boldsymbol{\omega},\mathbf{n}_{\mathbf{x}}\rangle^2,
\end{split}
\end{equation}
where $\mathrm{d}A(\mathbf{x})$ is an infinitesimal area around point $\mathbf{x}$ on the object surface $\mathcal{S}$, $\rho$ is the albedo, $\delta$ is the Dirac delta function, $c$ the speed of light, and $\boldsymbol{\omega}$ is a ray direction from $\mathbf{x}$ to $\mathbf{s}$. Time-independent effects such as $V$, the visibility function governed by scene geometry and sensor FoV, $f_r$ the bidirectional reflectance distribution function (BRDF), and foreshortening effects where $\mathbf{n}_{\mathbf{x}}$ is the surface normal at $\mathbf{x}$, are modeled as $G(\mathbf{x},\boldsymbol{\omega})$. 

\smallskip
\noindent\textbf{Sensor model.}
The sensor model accounts for laser and detector characteristics when converting a transient waveform into a transient histogram~\citeMain{hernandez2017computational}. Our sensor model considers the laser pulse, laser power, detector quantum efficiency, ambient photon flux, internal detector noise, pile-up effect, and time jitter. In practice, the \emph{laser pulse} is not a perfect impulse and, despite bandpass filters, the measured transient also captures some constant ambient light. To model this, we convolve $\tau(t)$~\footnote{We drop the subscript $\mathbf{s}$ hereafter for clarity.} with the laser's impulse response $g(t)$, scaled by $\phi^{scale}$ which absorbs laser power and quantum efficiency of the detector, and then offset its intensity by $\phi^{bkgd}$ which encapsulates ambient photon flux and internal detector noise:
\begin{equation}\label{eq:scale}
    \tilde{\tau}(t)=\phi^{scale}(\tau*g)(t)+\phi^{bkgd}.
\end{equation}
$\tilde{\tau}(t)$ is subsequently discretized into a histogram of Poisson rates $\mathbf{r}=[r_1,...,r_B]$ with $B$ bins. The probability $q_i$ of at least one photon falling inside the $i^{\mathrm{th}}$ bin is given by~\citeMain{coates1968correction}
\begin{equation}
    q_i=1-\exp(-r_i).
\end{equation}

In practice, SPADs aggregate photon counts over $C$ laser cycles, with only the first incident photon being detected in each cycle. This results in \emph{pile-up}, a nonlinear distortion of transients, leaving photons arriving at a later timestamp less likely to be detected~\citeMain{pediredla2018pileup}. Specifically, the probability $p_i$ of detecting a photon in the $i^{\mathrm{th}}$ bin in a cycle is given by~\citeMain{pediredla2018pileup}
\begin{equation}\label{eq:pileup}
    p_i=q_i\Pi_{k=1}^{i-1}(1-q_k).
\end{equation}
The photon counts $[h_1,...,h_B]$ in a transient histogram $\tilde{\mathbf{h}}$ follow a multinomial distribution:
\begin{equation}\label{eq:multinomial}
    [h_1,...,h_{B+1}]\sim\mathrm{Multinomial}(C, (p_1,..., p_{B+1})),
\end{equation}
where $p_{B+1}=1-\sum_{i=1}^B p_i$, and $h_{B+1}$ counts the number of cycles without detected photons. $\tilde{\mathbf{h}}$ is subsequently convolved with a discretized \emph{time jitter} kernel $\mathbf{s}$ to account for the temporal uncertainty of photon detection events, yielding the final histogram $\mathbf{h}$ measured by a SPAD detector:
\begin{equation}\label{eq:jitter}
    \mathbf{h}[b]=\Sigma_{k}\mathbf{h}[k]\mathbf{s}[b-k].
\end{equation}

We now present a reconstruction algorithm based on the analysis-by-synthesis principle to recover scene geometry from the transients. 
The input to this system is a distributed set of posed transients $\{\mathbf{h}_j\}_{j=1}^N$ captured of the scene.

\subsection{Neural Scene Reconstruction from SPADs}\label{sec:algorithm}

The reconstruction problem that we attempt to solve is extremely challenging. Unlike in conventional depth ranging where the depth of a scene point can be directly determined from the histogram via peak finding, a transient from our system represents the superposition of light reflected from numerous scene points, as illuminated by a diffuse laser, and is further contaminated by non-idealities of the detector (\eg, pile-up). The direct inversion of the signal is thus a highly ill-posed problem. Further, we cannot adapt methods from the NLOS imaging literature~\citeMain{o2018lct,lindell2019fk,xin2019theory} as they only support dense 2D scans, whereas our system uses a distributed, sparse and unstructured set of measurements.

To overcome these challenges, we resort to an analysis-by-synthesis approach based on differentiable rendering. Our approach allows flexible positioning of sensors and accurate modeling of histogram formation, thereby enabling high-quality reconstruction of scene geometry. We now describe our reconstruction algorithm in detail.

\smallskip
\noindent\textbf{Neural scene representation.}
Following NeuS~\citeMain{Wang2021neus}, we represent the scene geometry as a signed distance function (SDF), parameterized as a multi-layer perceptron (MLP) $f_{\theta}:\mathbb{R}^3\rightarrow\mathbb{R}$. $f_{\theta}$ maps the position-encoded (PE) $xyz$-coordinates of a point $\mathbf{x}$ to its signed distance $d$:
\begin{equation}
    d=f_{\theta}(\mathrm{PE}(\mathbf{x})).
\end{equation}
Compared to~\citeMain{Jungerman_ECCV_22,Sifferman2023unlocking}, this neural SDF allows our method to represent scene geometry beyond simple parametric shapes as the level set $\mathcal{S}=\{\mathbf{x}\in\mathbb{R}^3|f_{\theta}(\mathbf{x})=0\}$.

\smallskip
\noindent\textbf{Transient volume rendering.} The key idea behind our analysis-by-synthesis approach is to render $f_{\theta}$ into transients and compare them with those captured by our system. To adapt Equation~\ref{eq:path} for the rendering of $f_{\theta}$, we first rewrite it in angular integral form as
\begin{equation}\label{eq:angular}
\begin{split}
    \tau(t)=\int_{\Omega}&\frac{\rho}{\pi}\frac{V(\mathbf{x})\langle-\boldsymbol{\omega},\mathbf{n}_{\mathbf{x}}\rangle}{\|\mathbf{x}\|^2}
    \delta\left(\|\mathbf{x}\|_2-\frac{ct}{2}\right)\mathrm{d}\boldsymbol{\omega},
\end{split}
\end{equation}
where $\boldsymbol{\omega}$ are ray directions in the sensor FoV $\Omega$, and $\mathbf{x}$ the point where $\boldsymbol{\omega}$ intersects with the object surface $\mathcal{S}$ ($\infty$ if no intersection). For simplicity, we assume a learned spatially uniform albedo $\rho$ and Lambertian BRDF $f_r=1/\pi$~\footnote{This assumption may be relaxed to allow more expressive BRDF models such as the Phong reflection model~\citeMain{bui1975phong}.}. 

Inspired by NeRF~\citeMain{Mildenhall2020nerf} and NeuS~\citeMain{Wang2021neus}, we approximate Equation~\ref{eq:angular} via volume rendering to resolve surface discontinuities, enabling the optimization of $\theta$ via gradient descent:

\begin{equation}
    \hat{\tau}(t)=\int_{\Omega}\frac{\rho}{\pi}\frac{T^2(t)\sigma(\mathbf{p}(\boldsymbol{\omega},t))\langle-\boldsymbol{\omega},\mathbf{n}_{\mathbf{p}}\rangle}{\|\mathbf{p}(\boldsymbol{\omega},t)\|^2}\mathrm{d}\boldsymbol{\omega}.
\end{equation}
Here, $\mathbf{p}(\boldsymbol{\omega},t)=\nicefrac{ct}{2}\,\boldsymbol{\omega}$ are points along $\boldsymbol{\omega}$, the volume density $\sigma$ is a function of $f_{\theta}$ as in NeuS~\citeMain{Wang2021neus}, and the transmittance $T$ is given by
\begin{equation}
    T(t)=\exp{\left(-\int_0^t\sigma(\mathbf{p}(u))\mathrm{d}u\right)}.
\end{equation}

In practice, we discretize $\hat{\tau}(t)$ over the transient bin intervals $\{[t_i,t_{i+1})\}_{i=1}^B$ and work with the histogram $\hat{\boldsymbol{\tau}}=[\hat{\tau}_1,...\hat{\tau}_B]$, where
\begin{equation}\label{eq:volume}
    \hat{\tau}_i=\int_{\Omega}\frac{\rho}{\pi}\int_{t_i}^{t_{i+1}}\frac{T^2(t)\sigma(\mathbf{p}(\boldsymbol{\omega},t))\langle-\boldsymbol{\omega},\mathbf{n}_{\mathbf{p}}\rangle}{\|\mathbf{p}(\boldsymbol{\omega},t)\|^2}\mathrm{d}t\,\mathrm{d}\boldsymbol{\omega}.
\end{equation}

We estimate the intractable Equation~\ref{eq:volume} via Monte Carlo sampling of $\boldsymbol{\omega}$ and subsequently of $\mathbf{p}(\boldsymbol{\omega},t)$.

\smallskip
\noindent\textbf{Bilevel importance sampling.}
 Similar to~\citeMain{Mildenhall2020nerf,Wang2021neus}, the sampling of $\mathbf{p}(\boldsymbol{\omega},t)$ is weighted by a probability density function (PDF) over the equally sized bin intervals. This PDF is proportional to the per-bin weights $w_i$ given by
\begin{equation}
    w_i=\exp{\left(-\Sigma_{j=1}^{i-1}\sigma_j\Delta\right)}(1-\exp{(-\sigma_i\Delta})),
\end{equation}
where $\sigma_i$ is evaluated at the mid-point of the $i^{\mathrm{th}}$ bin, and $\Delta$ is the bin size in distance.

We extend this idea to the importance sampling of $\boldsymbol{\omega}$. Specifically, the sampling PDF over a uniform partitioning of FoV $\Omega$ is proportional to the cumulative weights $w^{(k)}$ over rays $\boldsymbol{\omega}^{(k)}$ drawn from each partition $k$:
\begin{equation}
    w^{(k)}=\Sigma_{i=1}^{B}w^{(k)}_i.
\end{equation}
Intuitively, this allows us to point more rays at high-density regions occupied by the object surface.

\smallskip
\noindent\textbf{Differentiable sensor modeling.}
Modeling sensor behavior is particularly important for our analysis-by-synthesis approach. This is because the synthesis targets $\boldsymbol{\tau}$ are not determined by the scene geometry alone but reflect the complex interplay of geometry with sensor non-idealities including pulse shape, pile-up and time jitter. To this end, we cascade $\hat{\boldsymbol{\tau}}$ to a \emph{differentiable sensor model} $\Gamma$ to simulate the transformation applied by the sensor to raw waveforms.

Specifically, $\Gamma$ closely follows the sensor model in Section~\ref{sec:model}; Equations~\ref{eq:scale}-\ref{eq:pileup} are differentiable and applied sequentially on $\hat{\boldsymbol{\tau}}$, yielding per-bin photon detection probabilities $\hat{\mathbf{p}}=[\hat{p}_1,...,\hat{p}_B]$. Instead of sampling photon counts using Equation~\ref{eq:multinomial}, we directly convolve $\hat{\mathbf{p}}$ with the jitter kernel as in Equation~\ref{eq:jitter}. This allows us to sidestep the non-differentiable sampling step while producing an unbiased estimate of the transient $\hat{\mathbf{h}}=\Gamma(\hat{\boldsymbol{\tau}})$ for loss evaluation.

\smallskip
\noindent\textbf{Loss functions.}
The optimization of $\theta$ is driven by three loss terms: a histogram reconstruction loss $\mathcal{L}_{hist}$ that minimizes the L1 distance between $\hat{\mathbf{h}}$ and $\mathbf{h}$, an Eikonal loss~\citeMain{gropp2020eikonal} $\mathcal{L}_{Eikonal}$ that regularizes the SDF, and a total variation regularizer~\citeMain{mu2022physics} $\mathcal{L}_{TV}$ that penalizes floaters in empty space. The combined loss function $\mathcal{L}$ is thus given by
\begin{equation}
    \mathcal{L}=\mathcal{L}_{hist} + \lambda_{Eikonal}\mathcal{L}_{Eikonal} + \lambda_{TV}\mathcal{L}_{TV},
\end{equation}
where $\lambda_{Eikonal}$ and $\lambda_{TV}$ are the respective loss weights.

\smallskip
\noindent\textbf{Comparison to NLOS-NeuS~\citeMain{fujimura2023nlos-neus}.} NLOS-NeuS is a concurrent work for scene reconstruction using SPAD sensors. Despite similarities in the imaging model and reconstruction algorithm, the two works differ substantially in application (direct line-of-sight \vs non-line-of-sight), laser characteristics (eye-safe diffuse laser \vs. high-energy beam laser), sensor quality and cost (commodity, low-cost \vs laboratory-grade, high-cost), transient modeling (with \vs without sensor non-idealities) and scan pattern (distributed and sparse \vs dense and structured 2D grid). We believe the two works complement each other and together unveil an exciting avenue toward single-photon 3D vision systems.

\section{Experiments}
\label{sec:experiments}

We demonstrate the effectiveness of our method for 3D geometric reconstruction of various objects in simulation, and in the real world with a low-cost SPAD. We provide qualitative and quantitative results for both settings. See the supplement for implementation details (\eg learning rates).


\begin{table*}[th]
\small
\centering
    \begin{tabular}{lrrrrrrrr}
        \toprule
        & \multicolumn{8}{c}{Chamfer Distance (mm) $\downarrow$} \\
        Method& Armadillo & Bear & Bunny & Digit & Einstein & Skull & Soap & Sphere \\
        \midrule
        Reprojection (Peak) & 54.29 & 40.36 & 34.95 & 55.85 & 43.25 & 48.90 & 51.71 & 51.07 \\
        Reprojection (Threshold) & 65.43 & 60.72 & 54.05 & 60.64 & 61.31 & 65.14 & 68.74 & 63.16 \\
        Space Carving & \underline{34.78} & \underline{24.53} & \underline{22.29} & \underline{45.44} & \underline{26.60} & \underline{25.49} & \underline{21.44} & \underline{25.47} \\
        Ours & \textbf{3.93} & \textbf{5.95} & \textbf{3.84} & \textbf{3.27} & \textbf{3.51} & \textbf{3.22} & \textbf{3.23} & \textbf{3.77} \\
        \bottomrule
    \end{tabular}\vspace{-0.5em}
    \caption{\textbf{Quantitative results on simulated data}. Our method more accurately recovers 3D shapes than baselines across $8$ objects.}\vspace{-0.5em}
    \label{tab:sim_quan}
\end{table*}

\begin{figure*}[ht]
    \centering
    \includegraphics[width=0.8\linewidth]{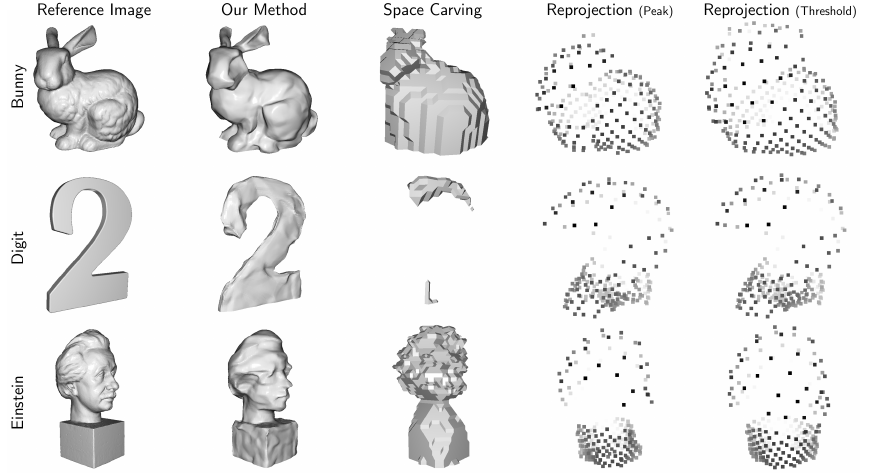}\vspace{-0.5em}
    \caption{\textbf{Qualitative results on simulated data}. Our method reconstructs dense and detailed 3D shapes. Space carving provides only hulls of a target shape, and is prone to carving away extra space when thin structures are present. Reprojection yields sparse points.}
    \label{fig:sim_qual}\vspace{-1.5em}
\end{figure*}

\smallskip
\noindent\textbf{Baselines.}
We compare our method to two baselines: reprojection~\citeMain{heide2018sub,gupta2019photon} and space carving~\citeMain{kutulakos2000space,tsai2017geometry}.

\emph{Reprojection}, also known as back-projection, reconstructs a scene as a point cloud and is the \emph{de facto} standard for depth ranging. We compare to two forms of reprojection. The peak method finds the distance $d$ corresponding to the histogram bin with the highest intensity. For a sensor at position $\mathbf{s}$ with an outwards pointing optical axis $\mathbf{u}$, a point is placed in the scene at position $\mathbf{s}+d\mathbf{u}$. The threshold method works in the same way but finds $d$ by locating the lowest-index bin with intensity above a threshold $t_p$. If no bin passes the threshold, no point is projected. We do not apply surface reconstruction to the generated point clouds, as it is often unreliable for poorly imaged scenes. See the supplement for further discussion.

\emph{Space carving} reconstructs a scene as a voxel grid. Like thresholded reprojection, it finds the distance $d$ corresponding to the lowest-index bin with intensity above a threshold $t_s$. All voxels in the sensor's FoV and nearer than $d$ are marked empty, along with voxels outside the FoV. Voxels in the FoV and further than $d$ are marked as occupied. The carved scene is the union of the occupied set for all sensors.

To ensure strong baselines for real-world experiments, we perform a brute-force search over $t_p$ and $t_s$ and choose values that minimize Chamfer distance over the entire real-world dataset. Space carving voxel size was set to $1.0\mathrm{cm}$.


\smallskip
\noindent\textbf{Evaluation protocol.}
Following NeuS~\citeMain{Wang2021neus}, we evaluate all methods using Chamfer distance. We report standard (two-way) Chamfer on simulated data. For real-world captures, we report Chamfer in both directions to evaluate the quality of reconstruction. Prior to Chamfer calculation, we convert ground-truth meshes and reconstructions from our method to point clouds by drawing $5$ million points uniformly at random on the mesh surface. For space carving, occupied voxels are converted to points if they touch unoccupied space, excluding the edge of the grid.

\begin{table*}[th]
\centering
\small 
    \begin{tabular}{
        l 
        @{\hspace{1em}}r@{\hspace{0.5em}}
        @{\hspace{0em}}r@{}
        @{\hspace{0em}}r@{}
        @{\hspace{1em}}r@{\hspace{0.5em}}
        @{\hspace{0em}}r@{}
        @{\hspace{0em}}r@{}
        @{\hspace{1em}}r@{\hspace{0.5em}}
        @{\hspace{0em}}r@{}
        @{\hspace{0em}}r@{}
        @{\hspace{1em}}r@{\hspace{0.5em}}
        @{\hspace{0em}}r@{}
        @{\hspace{0em}}r@{\hspace{0em}}
        @{\hspace{1em}}r@{\hspace{0.5em}}
        @{\hspace{0em}}r@{}
        @{\hspace{0em}}r@{\hspace{0em}}
    }
        \toprule
        & \multicolumn{15}{c}{Chamfer Distance (mm) $\downarrow$  \hspace{0.25em} \footnotesize{\textcolor{gray}{Rec $\rightarrow$ GT / GT $\rightarrow$ Rec}}} \\
        Method & \multicolumn{3}{c}{Big Box$^{*}$} & \multicolumn{3}{c}{Block$^{*}$} & \multicolumn{3}{c}{Pyramid$^{*}$} & \multicolumn{3}{c}{Toy Container$^{\dagger}$} & \multicolumn{3}{c}{Cereal Box$^{\dagger}$} \\
        \midrule
        Reprojection (Peak) & 77.4 & \footnotesize{\textcolor{gray}{24.9/}} & \footnotesize{\textcolor{gray}{52.5}} & 51.8 & \footnotesize{\textcolor{gray}{12.8/}} & \footnotesize{\textcolor{gray}{39.0}} & 94.7 & \footnotesize{\textcolor{gray}{17.5/}} & \footnotesize{\textcolor{gray}{77.1}} & \hspace{1em}71.0 & \footnotesize{\textcolor{gray}{24.9/}} & \footnotesize{\textcolor{gray}{46.0}} & \hspace{0.8em}49.3 & \footnotesize{\textcolor{gray}{17.3/}} & \footnotesize{\textcolor{gray}{31.9}}\hspace{0.8em} \\
        Reprojection (Threshold) & 67.5 & \footnotesize{\textcolor{gray}{14.8/}} & \footnotesize{\textcolor{gray}{52.7}} & 52.3 & \footnotesize{\textcolor{gray}{8.5/}} & \footnotesize{\textcolor{gray}{43.8}} & 75.4 & \footnotesize{\textcolor{gray}{\textbf{5.9}/}} & \footnotesize{\textcolor{gray}{69.5}} & 52.4 & \footnotesize{\textcolor{gray}{8.9/}} & \footnotesize{\textcolor{gray}{43.5}} & 51.6 & \footnotesize{\textcolor{gray}{19.2/}} & \footnotesize{\textcolor{gray}{32.4}}\hspace{0.8em} \\
        Space Carving & 67.9 & \footnotesize{\textcolor{gray}{35.1/}} & \footnotesize{\textcolor{gray}{32.8}} & 69.2 & \footnotesize{\textcolor{gray}{33.4/}} & \footnotesize{\textcolor{gray}{35.8}} & 80.1 & \footnotesize{\textcolor{gray}{39.5/}} & \footnotesize{\textcolor{gray}{40.6}} & 98.9 & \footnotesize{\textcolor{gray}{52.8/}} & \footnotesize{\textcolor{gray}{46.0}} & 44.1 & \footnotesize{\textcolor{gray}{24.4/}} & \footnotesize{\textcolor{gray}{19.6}}\hspace{0.8em} \\
        Ours & \textbf{12.5} & \footnotesize{\textcolor{gray}{\textbf{6.1}/}} & \footnotesize{\textcolor{gray}{\textbf{6.4}}} & \textbf{9.8} & \footnotesize{\textcolor{gray}{\textbf{5.6}/}} & \footnotesize{\textcolor{gray}{\textbf{4.2}}} & \textbf{18.4} & \footnotesize{\textcolor{gray}{9.0/}} & \footnotesize{\textcolor{gray}{\textbf{9.3}}} & \textbf{11.5} & \footnotesize{\textcolor{gray}{\textbf{5.8}/}} & \footnotesize{\textcolor{gray}{\textbf{5.6}}} & \textbf{16.3} & \footnotesize{\textcolor{gray}{\textbf{8.3}/}} & \footnotesize{\textcolor{gray}{\textbf{8.0}}}\hspace{0.8em} \\
        \bottomrule
    \end{tabular}\vspace{-0.5em}
    \caption{\textbf{Quantitative results on real-world captures}. Our method more accurately reconstructs real-world objects with homogeneous (*) and rich ($^{\dagger}$) texture. Reprojection yields sparse and unevenly distributed points, harming one-way Chamfer from GT to reconstruction.}\vspace{-1.5em}
    \label{tab:real_quantitative}
\end{table*}

\subsection{Simulated Experiments}
\label{subsec:simulated_experiments}

\noindent\textbf{Experiment setup.}
We simulate transients for eight scenes of varying complexity using the image formation model in Section~\ref{sec:model}. The objects are centered on the ground plane ($z=0$) with the largest dimension $\approx0.3\mathrm{m}$. Sensors with a conical FoV are uniformly distributed on a hemisphere at the origin with a radius of $0.5\mathrm{m}$, and are all pointed at the origin. In our simulation, $N=256$, $B=256$, $\Delta=5\mathrm{mm}$, $\mathrm{FoV}=30^{\circ}$, $\phi^{scale}=1$, $\phi^{bkgd}=0.001$, $C=5000$ and $\rho=0.8$. The laser pulse, $g$, has a full-width-at-half-maximum (FWHM) of $50\mathrm{ps}$, and $\mathbf{s}$ is a tabulated PDF obtained from experiments~\citeMain{hernandez2017computational}. The sensor parameters are deliberately chosen to reflect the characteristics of low-cost sensors. See the supplement for a sensitivity study on how these parameters affect the quality of reconstruction.

\smallskip
\noindent\textbf{Results.}
Table~\ref{tab:sim_quan} summarizes the quantitative results of all methods. Our method achieves an average Chamfer distance of $<5\mathrm{mm}$, an order of magnitude lower than all baselines. A key reason is that the baselines only use depth information from a single histogram bin, whereas our method makes effective use of the entire waveform, which contains rich geometry cues about a large scene patch.

We provide visualizations of our results in Figure~\ref{fig:sim_qual}. Our method recovers global scene structure as well as local geometry details. In contrast, reprojection yields sparse point clouds without sufficient coverage of the scene. While space carving produces dense reconstructions, the occupancy grid only represents an envelope of the scene, leaving it difficult to recognize the precise shape of an object.

\begin{figure}[t]
    \centering
    \includegraphics[width=1.0\linewidth]{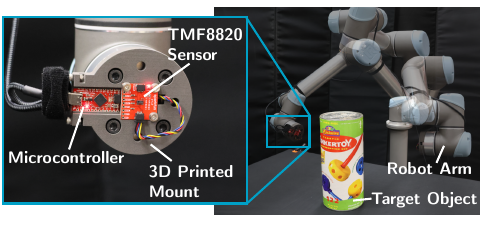}\vspace{-1em}
    \caption{To capture real-world data from a wide set of viewpoints, we mount the TMF8820 proximity sensor to a robot arm. Forward kinematics of the robot are used to gather sensor pose.}
    \label{fig:sensor_and_robot}\vspace{-1.5em}
\end{figure}

\begin{figure*}[ht]
    \centering
    \includegraphics[width=0.9\linewidth]{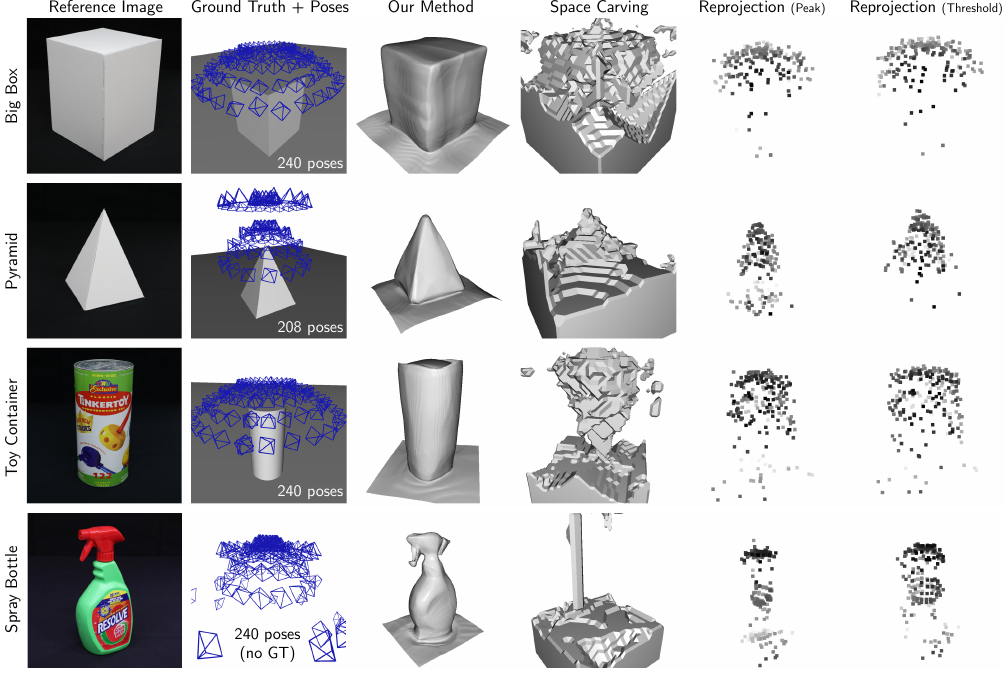}\vspace{-0.5em}
    \caption{\textbf{Qualitative results on real-world captures}. Our method again attains the highest reconstruction quality. Poses in column two are subsampled by a factor of two for clarity. See supplement for additional qualitative results.}\vspace{-1.5em}
    \label{fig:qualitative_real}
\end{figure*}

\subsection{Hardware Prototype}
\label{subsec:considerations_real_hardware}

We use the SPAD-based AMS TMF8820 proximity sensor~\citeMain{TMF8820}, which retails for \$10 USD. We connect the sensor to a microcontroller via I$^2$C and use the AMS-provided driver to extract transient histograms.

The sensor contains a total of 216 SPADs, which are pooled onboard the sensor into $3\times3$ zones, each of which images a different FoV. The sensor captures one transient histogram for each zone. We pool histograms from all zones, which is equivalent to capturing one wide-FoV histogram per-measurement as SPADs do not suffer from readout noise~\citeMain{zappa2007principles}. In doing so, we avoid inter-histogram interference previously observed by~\citeMain{Sifferman2023unlocking} and avoid the need to model individual fields-of-view of the sensor, which we empirically observed to have soft and poorly specified boundaries. We slightly modify our method to accommodate the AMS TMF8820 sensor used in real-world experiments.

\smallskip
\noindent\textbf{Laser Impulse.}
The laser impulse response of the TMF8820 is not Gaussian and varies slightly between measurements. Fortunately, the sensor captures the shape of its laser impulse for each measurement in a ``reference histogram''. We record this histogram for each measurement and incorporate it into our forward model by cross-correlating the idealized scene response with this recorded reference histogram. We observe that the bin size $\Delta_r$ of the reference histogram is smaller than the bin size $\Delta$ of the transient histograms captured by the sensor. To account for this, we scale the reference histogram in the temporal dimension by a factor $\Delta_r / \Delta$ before cross-correlation. Further, we find that it is necessary to temporally shift the reference histogram by a fixed amount $\phi^{delay}$ before correlation.

To calibrate the parameters $\Delta$, $\Delta_r$, and $\phi^{delay}$, we perform the one-off intrinsic calibration procedure separately introduced by \citeMain{Sifferman2023unlocking}. The TMF8820 sensor is pointed at a planar surface from a range of known distances and angles-of-incidence. A differentiable render-and-compare method is used to optimize for the unknown sensor intrinsic parameters given known planar geometry. 

\smallskip
\noindent\textbf{Pile-up Correction.}
While our forward model assumes that the target transients exhibit nonlinear distortion due to pile-up, the TMF8820 sensor performs pile-up correction on-sensor, and it cannot be disabled. To accommodate this, we incorporate the differentiable Coates' correction~\citeMain{coates1968correction} as a final step in the forward model.

\smallskip
\noindent\textbf{Other Sensors.} Our design can be easily applied to any SPAD with a co-located diffuse illumination source, including other low-cost sensors~\citeMain{VL53L8CH}, high-end setups \citeMain{Jungerman_ECCV_22, Nishimura2020disambiguating}, and things in between. The only modification necessary is the calibration of sensor intrinsics (\eg, using~\citeMain{Sifferman2023unlocking}). 


\subsection{Real-world Experiments}
\label{subsec:real_world_experiments}
\noindent\textbf{Experiment Setup.}
We capture a real-world tabletop dataset of eight objects\footnote{See supplement for full results over the entire dataset.} of varying geometry and texture. To capture many posed views of the target object, we attach the sensor to a Universal Robots UR5 robot arm. We program the arm to automatically move to a set of poses and record sensor measurements at each pose. To obtain sensor poses, we use the forward kinematics of the robot, which are accurate to $\pm0.5$mm \citeMain{pollak2020measurement}. Each object is captured from between 128 and 240 viewpoints. Five of the objects are simple geometric primitives, for which we manually generate ground-truth meshes based on the dimensions of the target object and measurements of its position from the robot's forward kinematics. Meshes are trimmed to an axis-aligned bounding box 16cm larger than the target object in each dimension before the Chamfer distance calculation.

\begin{figure}
    \centering
    \includegraphics[]{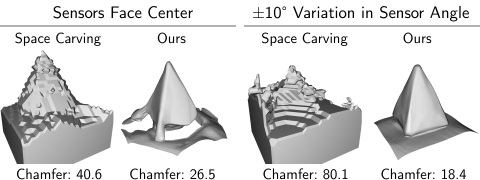}\vspace{-0.5em}
    \caption{Space carving performs more poorly when sensor poses include some variation in target point, while our system takes advantage of the increased view diversity and coverage.}\vspace{-1em}
    \label{fig:randomness}
\end{figure}

\smallskip
\noindent\textbf{Results.}
As seen in Table~\ref{tab:real_quantitative}, our method outperforms all baselines by a wide margin as measured by two-way Chamfer distance. While reprojection is at times competitive in one-way distance from reconstruction to ground truth, it performs poorly in the opposite direction due to the sparse and unevenly distributed point cloud generated, as visualized in Figure~\ref{fig:qualitative_real}. While space carving outperforms reprojection on simulated data under a highly structured sensor pose distribution (\ie all sensors are facing the center of the object), it yields poor results on real-world scenes, in which we vary sensor orientation by $\pm10^{\circ}$ to emulate real-world capture conditions and increase coverage. By contrast, our method benefits from the more varied sensor poses as is shown in Figure~\ref{fig:randomness}.

\begin{figure}
    \centering
    \includegraphics[]{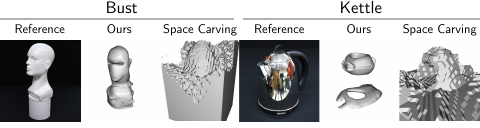}\vspace{-0.5em}
    \caption{Because our reconstruction method assumes a Lambertian surface, it fails to reconstruct highly specular scenes, such as a glossy white bust (left) or mirror-finish kettle (right).}\vspace{-2.0em}
    \label{fig:failure}
\end{figure}

Further, our method is surprisingly robust to violation of assumptions made about surface reflectance; it successfully reconstructs non-Lambertian objects with rich texture despite assuming Lambertian BRDF with a spatially uniform albedo. These include both simple shapes (Toy Container in Figure~\ref{fig:qualitative_real} and Cereal Box) and challenging objects with complex geometry (Spray Bottle in Figure~\ref{fig:qualitative_real}). We hypothesize that the overlapping FoVs of distributed sensors help constrain the optimization of our model and encourage a plausible reconstruction that best explains all transients. Our strong results on simulated and real-world data validate our modeling approach and demonstrate a single-photon 3D vision system for real-world scene reconstruction.

\section{Discussion and Future Work}
\label{sec:discussion}

Despite assuming spatial uniform reflectance and albedo, our method is robust to rich textures (Fig.~\ref{fig:qualitative_real}) and compares favorably to baseline methods for reconstructing challenging scenes with high specularities (Fig.~\ref{fig:failure}). Future work will investigate recovering spatially varying reflectance or incremental learning of geometry as new measurements become available~\citeMain{OrtiziSDF2022, Sucar2021imap}, enabling applications like real-time mapping and SLAM. Our method may be particularly relevant in applications such as robotics and wearable computing, where the small size, low power requirements, and robust hardware of proximity sensors are very valuable.

\vspace{0.3em}
\noindent{\small \textbf{Acknowledgements:} This work was supported by Los Alamos National Lab and the Department of Energy, Wisconsin Alumni Research Foundation, NSF CAREER Award 1943149, NSF grant CNS-2107060, and a SONY Faculty Innovation Award. Gleicher and Gupta hold concurrent appointments at Amazon and Cruise, respectively. This work is not associated with either company.
}

{
    \bibliographystyleMain{ieeenat_fullname}
    \bibliographyMain{main}
}

\setcounter{page}{1}
\setcounter{figure}{0}
\setcounter{table}{0}
\setcounter{section}{0}
\setcounter{equation}{0}
\renewcommand{\thefigure}{\Alph{figure}}
\renewcommand{\thesection}{\Alph{section}}
\renewcommand{\thetable}{\Alph{table}}
\renewcommand{\theequation}{\Alph{equation}}
\maketitlesupplementary


In this supplement, we describe (1) The implementation details of our reconstruction pipeline (Section~\ref{sec:implementation}); (2) a sensitivity analysis to understand the robustness of our method w.r.t.\ imaging parameters (Section~\ref{sec:appendix_sensitivity}); (3) A discussion of the difficulty of point cloud surface reconstruction (Section~\ref{sec:point_cloud_recon}); (4) additional qualitative results on simulated data and real world captures (Section~\ref{sec:appendix_results}); and (5) further discussion of our work (Section~\ref{sec:appendix_discussion}). We also include three short videos, which provide animated $360^{\circ}$ views of qualitative reconstruction results. The videos can also be viewed on the project web page: \url{https://cpsiff.github.io/towards_3d_vision/}.

For sections, figures and equations, we use numbers (\eg, Sec.\ 1) to refer to the main paper and capital letters (\eg, Sec.\ A) to refer to this supplement.

\section{Implementation Details}
\label{sec:implementation}
For both simulated and real world experiments, we use an $8$-layer MLP with $256$ hidden units as our SDF, $f_{\theta}$, and initialize it as a sphere, centered at the origin with radius $0.3$m, using geometric initialization~\citeSupp{atzmon2020sal_}. For each transient, we sample $256$ rays $\boldsymbol{\omega}$ over $\Omega$ and sample $256$ points per ray. We set $\lambda_{Eikonal}$ to $0.1$ across all experiments and set $\lambda_{TV}$ to $0$ and $0.01$ respectively for the simulated and real-world experiments. We train $f_{\theta}$ for $300\mathrm{K}$ steps using Adam~\citeSupp{kingma2014adam_} with a mini-batch size $2$, a learning rate $0.0005$, and cosine decay. The learned SDFs are converted to meshes using Marching Cubes~\citeSupp{lorensen1998marching_}.

\section{Sensitivity Analysis} 
\label{sec:appendix_sensitivity}

\begin{figure*}
    \centering
    \includegraphics[width=\textwidth]{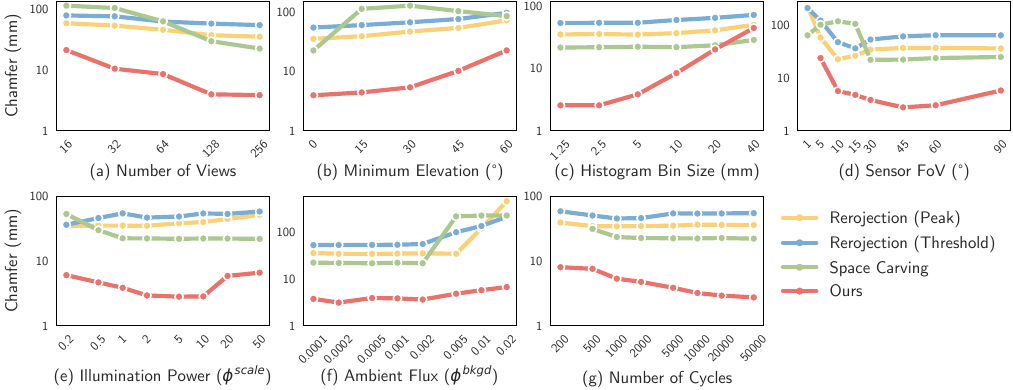}
    \caption{\textbf{Sensitivity analysis of our method compared to baselines across a range of imaging parameters.} In almost every case, our method outperforms baseline methods on Chamfer distance. Missing datapoint in (d) indicates that our method failed to converge. Illumination power (e) is unit-less as it also absorbs factors like quantum efficiency and does not map directly to any real world parameter.}\label{fig:sensitivity}
\end{figure*}

We perform extensive experiments to understand the robustness of our method in comparison to baselines under varying sensor parameters in simulation. All experiments are based on the Bunny scene and the parameters are varied one at a time while other parameters remain fixed at the base condition (as described in Section 4 of the main paper). To ensure strong baselines for every sensor configuration, we calibrate the thresholds $t_p$ and $t_s$ for the projection (threshold) and space carving baselines respectively per reconstruction. We perform a brute force search over possible thresholds and report the best Chamfer achieved. As this amounts to calibrating on the test set, the numbers reported represent the best possible performance of the baseline methods on the given data. The results of this sensitivity analysis are presented in Figure~\ref{fig:sensitivity}. In what follows we discuss some of the main findings.

\medskip
\noindent\textbf{Sensor Placement.} We study two key parameters that control sensor placement: the number of views and the minimum elevation angle at which the sensors are placed. Our method consistently outperforms all baselines in Chamfer distance by an order of magnitude across a broad range of parameter choices. In particular, our method readily supports as few as $128$ views above a considerably large elevation angle of $30^{\circ}$ without harming reconstruction quality. This robust gain in performance confirms that our method takes advantage of broad-band signal in transients not exploited by the baseline methods.

\medskip
\noindent\textbf{Temporal Resolution.} Our system takes advantage of the temporal information in transient histograms, and therefore benefits when that information is present at a high resolution. Because of this, our method outperforms baselines by a very wide margin at a small bin size, but the margin vanishes as bins become wider than $2\mathrm{cm}$ (equivalently $66\mathrm{ps}$), because decomposing the temporal signal becomes impractical beyond this limit. Fortunately, today's commodity SPADs operate at a smaller bin size ($\sim40\mathrm{ps}$). Baseline methods show no performance gain at small bin sizes, as they do not take advantage of the temporal resolution.

\medskip
\noindent\textbf{Angular Resolution.} Our system resolves spatial resolution from wide-FoV sensors by taking advantage of the time dimension. In this regime, the optimal sensor field-of-view size is not obvious: a smaller FoV means more highly constrained geometry, as each histogram images a smaller region, but too small of a field-of-view means a lack of coverage and under-constrained geometry. We find that an angular resolution in the $30^{\circ}$ to $60^{\circ}$ range is optimal for reconstructing 3D geometry with our method on the bunny scene. Reprojection based methods benefit more from a smaller field-of-view, while space carving performs best with a wider field-of-view so that space is sufficiently carved away. In every case, our method outperforms baselines by a wide margin.

\medskip
\noindent\textbf{Signal-to-noise ratio (SNR).} We consider three parameters that jointly impact SNR: illumination power, ambient flux, and number of illumination cycles. Our method again outperforms all baselines by a significant margin across all test conditions. Notably, the baselines fail or perform considerably worse under high ambient flux, as signal photons are blocked by background photons due to pile-up. By contrast, our method is robust against a broad range of ambient flux levels, as we model the effects of ambient flux directly.

\begin{figure}
    \centering
    \includegraphics[]{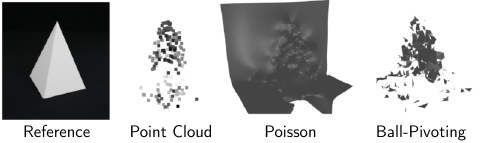}
    \caption{Off-the-shelf algorithms for surface reconstruction (Poisson Surface Reconstruction \protect\citeSupp{kazhdan2006poisson_} and The Ball-Pivoting Algorithm \protect\citeSupp{Bernardini1999ball_}) do not perform well on point clouds generated by reprojection.}
    \label{fig:supp_point_cloud_reconstruction}
    \vspace{-1em}
\end{figure}

\section{Point Cloud Surface Reconstruction}
\label{sec:point_cloud_recon}
We do not apply surface reconstruction to the point clouds computed by reprojection because off-the-shelf reconstruction techniques do not perform reliably on the generated point clouds, as shown in Figure~\ref{fig:supp_point_cloud_reconstruction}. Additionally, calculating Chamfer distance to the computed point cloud ensures that we are capturing the efficacy of reprojection rather than a given surface reconstruction method. We include one-way reconstruction to ground truth Chamfer distance in Table 2 to provide a metric which does not penalize the sparsity of point clouds produced by reprojection.

\section{Additional Qualitative Results}\label{sec:appendix_results}

\begin{figure*}
    \centering
    \includegraphics[width=\textwidth]{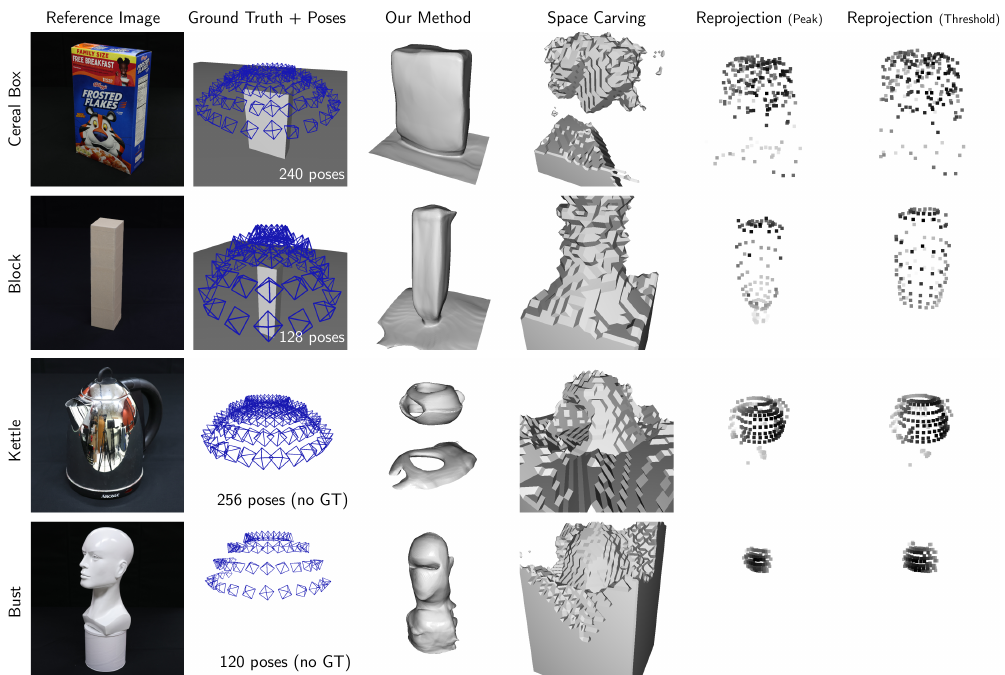}
    \caption{\textbf{Additional qualitative results on real-world captures.} Our method achieves the highest reconstruction quality. Poses in column two are subsampled by a factor of two for clarity.}\label{supp:real}
\end{figure*}

\begin{figure*}
    \centering
    \includegraphics[width=\textwidth]{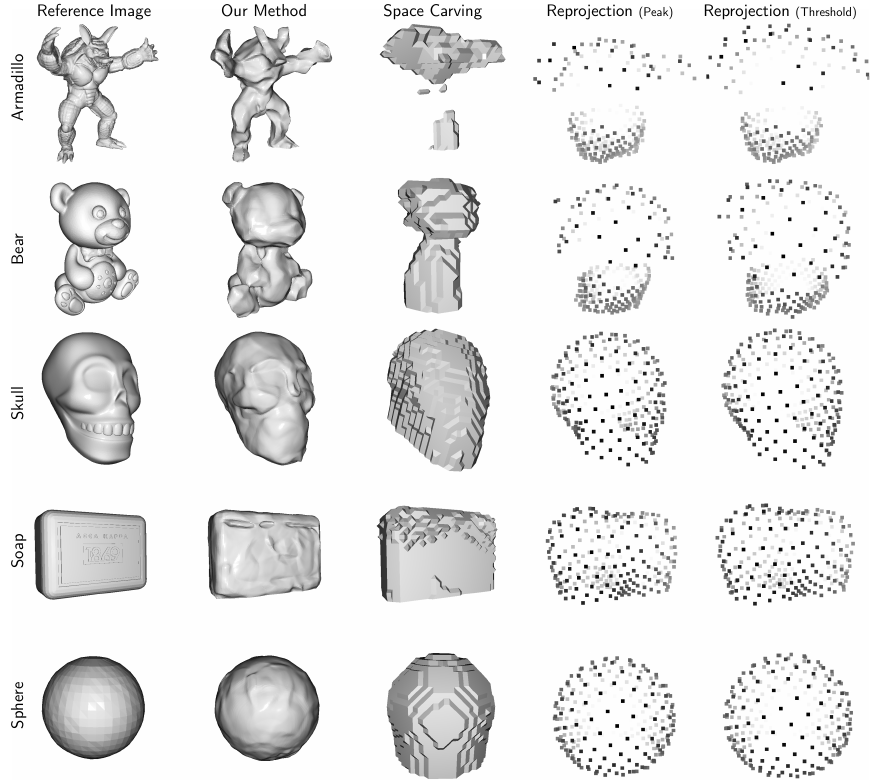}
    \caption{\textbf{Additional qualitative results on simulated data.} Our method achieves the highest reconstruction quality. Space carving captures an envelope of the shape, and may carve away occupied areas in concave shapes (\eg Armadillo). Reprojection gives a sparse reconstruction of convex shapes, (\eg skull, soap, sphere), the scale of which may be distorted due to biases introduced by the wide field-of-view of the sensor.}\label{supp:sim}
\end{figure*}

\noindent\textbf{More reconstruction results.} We present additional qualitative results on simulated data (Figure~\ref{supp:sim}) and real-world captures (Figure~\ref{supp:real}). These results were omitted from the main paper due to lack of space. 

\medskip
\noindent\textbf{Surface normal visualization.} Moving beyond the 3D shapes, we further examine the surface normal of our reconstructed 3D objects. 
The surface normal of a point $\mathbf{x}$ on the reconstructed mesh is estimated as
\begin{equation}
    \tilde{\mathbf{n}}_{\mathbf{x}}=\frac{\nabla_{\mathbf{x}}(f_{\theta}(\mathrm{PE(\mathbf{x})}))}{\|\nabla_{\mathbf{x}}(f_{\theta}(\mathrm{PE(\mathbf{x})}))\|},
\end{equation}
where $f_{\theta}$ is the learned SDF and PE denotes the positional encoding function. The error $e_{\mathbf{x}}$ w.r.t.\ the ground-truth normal $\mathbf{n}_{\mathbf{x}}$ is given by
\begin{equation}
    e_{\mathbf{x}}=|\langle\mathbf{n}_{\mathbf{x}}, \tilde{\mathbf{n}}_{\mathbf{x}}\rangle|.
\end{equation}

\begin{figure*}[t]
    \centering
    \includegraphics[width=\textwidth]{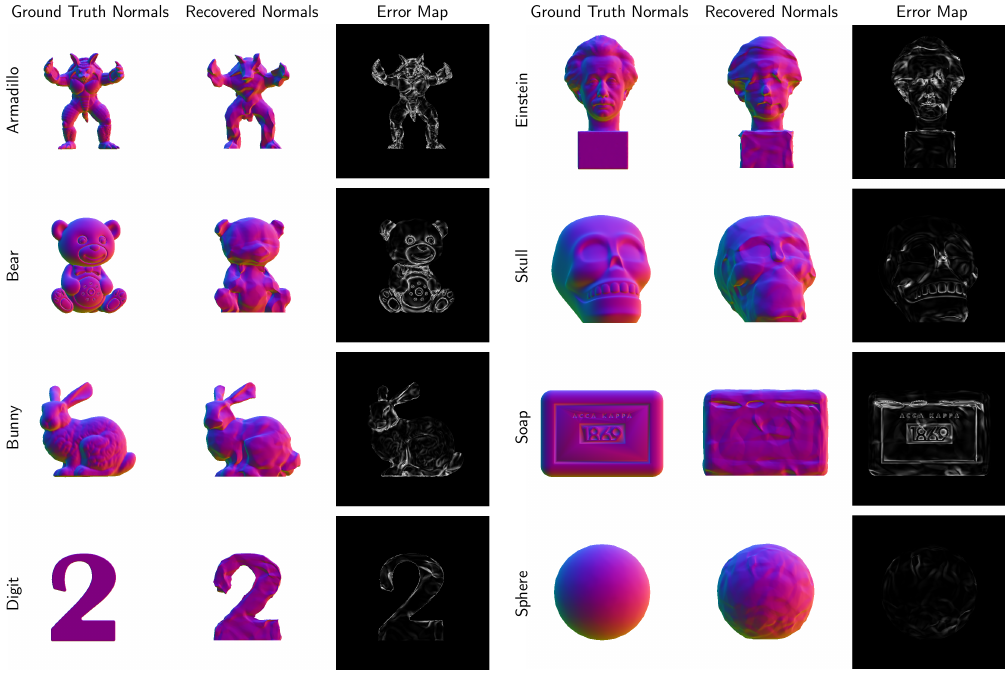}
    \caption{\textbf{Visualizations of surface normals for simulated data.} Our method correctly estimates surface normals in flat regions. Error mainly occurs at edges and depth discontinuities. We hypothesize that sensors with higher temporal and spatial resolution are needed to detect rapid changes in surface normals.}\label{supp:normal}
\end{figure*}

\medskip
\noindent\textbf{Surface normal results.} We provide visualizations of surface normals for simulated data in Figure~\ref{supp:normal}.  Our method can successfully recover smoothly varying normals. Error typically occurs at edges and depth discontinuities with fast-changing normals. We hypothesize that sensors with higher temporal and spatial resolution are needed for more accurate surface normal reconstruction.

\section{Further Discussion}\label{sec:appendix_discussion}

\noindent\textbf{Beyond Lambertian objects.} Our method assumes a spatially uniform Lambertian BRDF, but in practice can effectively reconstruct objects with spatially varying albedo and slightly glossy appearance (\eg the spray bottle). In theory, our method can easily be adapted to incorporate a parametric lighting model. Recovery of the parameters of such a model are likely possible because, by sharing information among many observations, the BRDF is effectively sampled at many incident and exitant angles. An intriguing direction for future work is investigating which BRDF parameterizations can be recovered with our imaging setup, and the effect of the reflectance model on reconstruction quality. We suspect that a non-parametric NeRF-like BRDF would not be suitable as it does not sufficiently constrain the optimization. A parametric lighting model, \eg Phong~\citeSupp{bui1975phong_} or Oren-Nayar~\citeSupp{oren1994oren-nayar_} may appropriately constrain the optimization while allowing the model to learn a more accurate scene representation.

\medskip
\noindent \textbf{Runtime efficiency.} Our method takes on the order of hours to reconstruct a scene, making it unsuitable for real-time applications in its current state. Future work should investigate ways to speed up forward rendering and model training. Improved importance sampling would likely yield modest improvements in convergence time. Another option is to render only summary statistics of the histogram (\eg mean, peak locations or widths) rather than the entire histogram, which would likely be faster to render at the expense of yielding a lower-quality reconstruction.


\medskip
\noindent \textbf{Sensor pose.} In this work, we used an industrial robot arm to gather posed sensor measurements. We chose this modality as it is guaranteed to provide highly accurate sensor poses, and allows control over precise sensor placement. For applications like wearable computing and distributed sensing for robotics, camera poses might be pre-calibrated and remain fixed relative to each other during operation. Alternatively, the low-cost single-photon camera could be combined with a sensor-based localization system (\eg, an IMU based~\citeSupp{yi2007imu_} or a camera based~\citeSupp{mur2015orb_} system) to recover camera pose, a setup which is standard in related works~\citeSupp{Mildenhall2020nerf_, OrtiziSDF2022_}. Such a capture setup would allow capture of more organic and large scale scenes, which more closely mimic the potential use cases of the sensor (\eg on mobile robots and drones).


\medskip
\noindent \textbf{Comparison to other 3D imaging modalities.}
Our work provides a low-cost 3D imaging system using single-photon cameras. We provide detailed comparisons between our method and baseline methods, but do not compare our reconstructions to those gathered from other 3D modalities, such as continuous wave time-of-flight \citeSupp{attal2021torf} or LiDAR \citeSupp{Huang2023nfl}. Future work should provide a comparison to these other modalities to provide insights into the niche (in terms of accuracy, size, power, etc.) filled by each.

\medskip
\noindent \textbf{Comparison to a NeRF with a miniature RGB camera.} Using multiple calibrated RGB cameras, NeRF-like approaches can leverage photometric cues (\eg correspondence and shading) to recover 3D shapes from a distributed set of cameras. \emph{Conceptually}, NeRF-like approaches fall short under non-ideal lighting (\eg low-light) or insufficient correspondence (\eg textureless objects), while our method with active lighting and depth remains effective. With sufficient light and distinct texture patterns, NeRF-like approaches on RGB images will yield a higher quality reconstruction than our method, due to significantly increased data rates. Without compression, a VGA-resolution image with 8-bit color channels contains 7Mbits of information, while an image (\ie, transient histogram) from our sensor contains 3Kbits of information. With $2400\times$ the information per-view, it is not surprising that NeRF with RGB cameras could outperform our method given the same number of input views. \emph{Practically}, existing low-cost RGB cameras are larger and less power efficient than the SPAD sensors that we utilize. While miniature RGB cameras do exist (\eg, those used for endoscopy), they are $>$20$\times$ the cost of a SPAD.

\medskip
\noindent \textbf{Commodity sensors.} One challenge for future work is a lack of hardware support for measurement and use of transient histograms. Very few low-cost sensors allow access to transient histograms, and those that do often perform pre-processing that is proprietary or undocumented. Additionally, most sensors are equipped with very low-bandwidth I$^2$C interfaces, limiting their effective FPS. We hope that manufacturers will see value in users having access to transient histogram data and support the use of this data with documentation, low-level access, and high-bandwidth interfaces in the future.

\medskip
\noindent \textbf{Ethical concerns.} Our work presents a new method for imaging 3D objects with low-cost single-photon cameras. We do not anticipate major ethical concerns.

\clearpage
\clearpage

{
    \bibliographystyleSupp{ieeenat_fullname}
    \bibliographySupp{main}
}

\end{document}